\pdfoutput=1

\documentclass[11pt]{article}

\usepackage[final]{acl}

\usepackage{times}
\usepackage{latexsym}

\usepackage[T1]{fontenc}

\usepackage[utf8]{inputenc}

\usepackage{microtype}

\usepackage{inconsolata}

\usepackage{tabularx}

\usepackage{url}
\usepackage{graphicx}
\usepackage{amsmath}
\usepackage{multirow}
\usepackage{booktabs}
\usepackage{listings}
\usepackage{wasysym}
\usepackage{amssymb}
\usepackage{mdframed}
\usepackage{authblk}
\usepackage{xcolor}
\usepackage{tcolorbox}
\tcbuselibrary{breakable}

\usepackage{enumitem}
\usepackage{graphicx}

\title{BacktrackAgent: Enhancing GUI Agent with Error Detection and Backtracking Mechanism}

\author{\bf Qinzhuo Wu, Pengzhi Gao, Wei Liu, Jian Luan \\
  MiLM Plus, Xiaomi Inc \\
  \texttt{\{wuqinzhuo, gaopengzhi, liuwei40, luanjian\}@xiaomi.com} \\}

\begin{document}
\maketitle
\begin{abstract}

Graphical User Interface (GUI) agents have gained substantial attention due to their impressive capabilities to complete tasks through multiple interactions within GUI environments. However, existing agents primarily focus on enhancing the accuracy of individual actions and often lack effective mechanisms for detecting and recovering from errors. To address these shortcomings, we propose the BacktrackAgent, a robust framework that incorporates a backtracking mechanism to improve task completion efficiency. BacktrackAgent includes verifier, judger, and reflector components as modules for error detection and recovery, while also applying judgment rewards to further enhance the agent's performance. Additionally, we develop a training dataset specifically designed for the backtracking mechanism, which considers the outcome pages after action executions. Experimental results show that BacktrackAgent has achieved performance improvements in both task success rate and step accuracy on Mobile3M and Auto-UI benchmarks. 
Our data and code will be released upon acceptance.


\end{abstract}

\section{Introduction}

Graphical User Interface (GUI) agents \cite{hong2024cogagentvisuallanguagemodel,ma-etal-2024-coco} have demonstrated remarkable capabilities to perform tasks within digital environments. Early advancements \cite{zhang2023appagentmultimodalagentssmartphone,zhang-etal-2024-android,yan2023gpt4vwonderlandlargemultimodal} were primarily based on general Vision-Language Models (VLM) such as GPT-4V and GPT-4o \cite{openai2023gpt4}. Since then, numerous GUI agent-specific datasets and models \cite{rawles2023androidwildlargescaledataset,ijcai2024p339,10.1007/978-3-031-73039-9_14,chai2024amexandroidmultiannotationexpo} have been developed. These agents are specifically designed to handle tasks involving graphical elements like buttons, text boxes, and images. By utilizing advanced perception and reasoning capabilities, these agents have the potential to transform task automation, improve accessibility, and optimize workflows across various applications.

\begin{figure}[!t]
\setlength{\abovecaptionskip}{0.2cm}
\setlength{\belowcaptionskip}{-0.3cm}
\centering
\includegraphics[width=0.85\columnwidth]{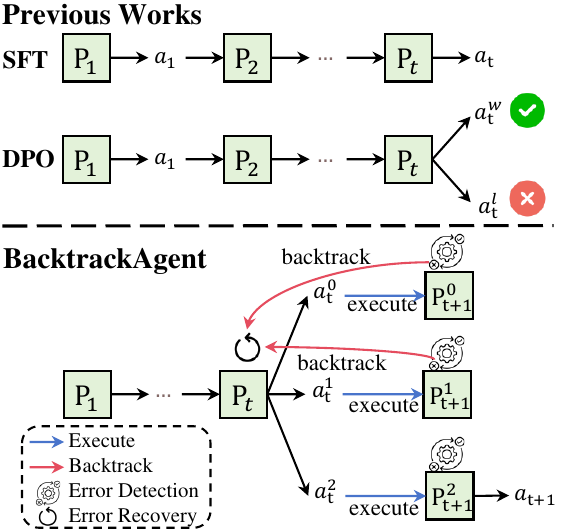}
\caption{Previous works often struggle to recover from errors, whereas BacktrackAgent utilizes a backtracking mechanism to recover from erroneous pages.}
\label{figure1-1}
\end{figure}

\begin{figure*}[!t]

\setlength{\abovecaptionskip}{0.2cm}
\setlength{\belowcaptionskip}{-0.3cm}
\centering
\includegraphics[width=0.9\textwidth]{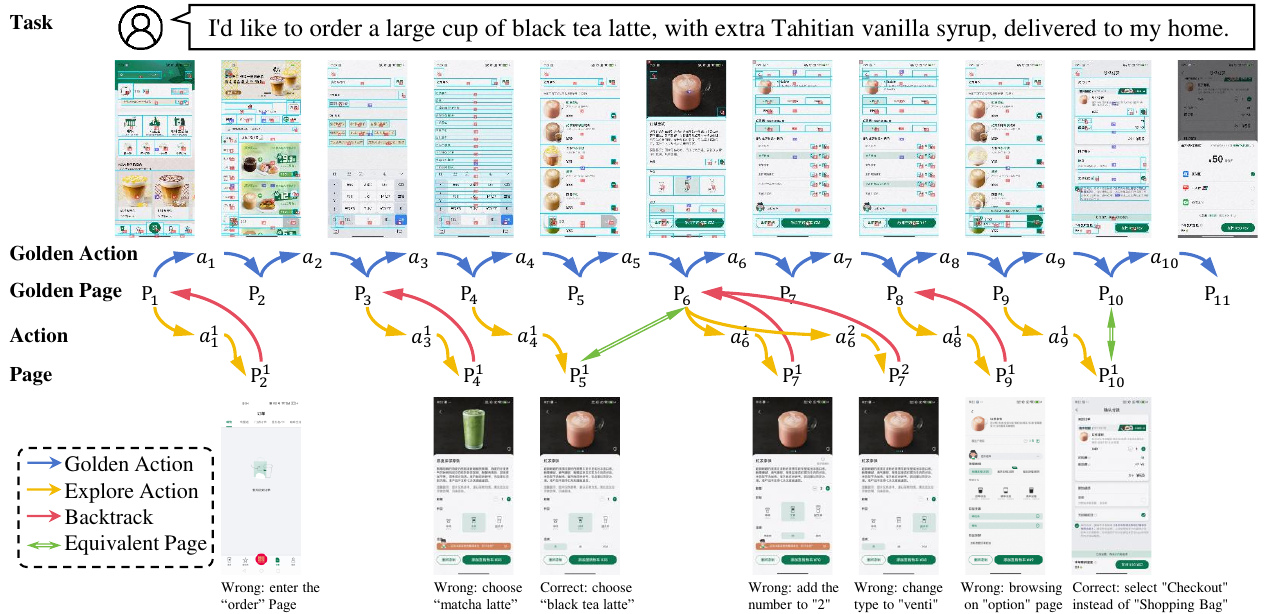}
\caption{A ten-step GUI trajectory for ordering coffee. The red arrow indicates that the current page is identified as an error page, requiring a backtrack to the previous page in order to regenerate the necessary action. Action $a_1$ is an abbreviation for "click(delivery,[375,740][704,1032])". The detailed information is summarized in Figure \ref{figure2-2}.}
\label{figure2-1}
\end{figure*}

Current GUI agents face several challenges when completing tasks, as they primarily focus on achieving single-step accuracy and struggle to recover from errors. As shown in Figure \ref{figure2-1}, a task may require more than ten actions to complete, and one incorrect action can result in the failure of the entire task. Most studies rely on supervised fine-tuning (SFT) using annotated page navigation datasets, which replicate successful cases while neglecting the understanding of error cases. Some studies based on preference optimization, such as DigiRL \cite{bai2024digirltraininginthewilddevicecontrol} and DistRL \cite{wang2024distrlasynchronousdistributedreinforcement}, generate numerous negative examples that are paired with positive examples, as illustrated by ($a^w_t, a^l_t$) in Figure \ref{figure1-1}. These methods encourage generated actions to avoid negative examples, aligning them with desired sampling preferences. However, preference optimization-based methods depend heavily on the quality and sufficiency of the sampled data. They do not consider the outcomes of executing actions, making it difficult to determine whether the current page deviates from the task, as well as to recover from any errors.

To address this issue, we propose BacktrackAgent, a framework designed to incorporate a backtracking mechanism that enhances task completion. BacktrackAgent consists of four components: generator, verifier, judger, and reflector. The generator creates and executes actions based on the current task and GUI environment. The verifier and judger act as error detection modules, determining whether the current state requires backtracking. The reflector functions as an error recovery module, refining the actions based on the judgments and guiding the agent back to a state that is most likely to lead to successful task completion. The rewards from the verifier and judger are utilized to further improve the agent's capabilities. The contribution of this paper can be summarized as follows:
\begin{itemize}[itemsep=2pt]
\item We propose BacktrackAgent, a framework that integrates a backtracking mechanism, which employs a verifier and a judger as error detection modules, along with a reflector acting as the error recovery module.
\item We construct judgment and reflection datasets based on the Mobile3M \cite{wu-etal-2024-mobilevlm} and Auto-UI \cite{zhang-zhang-2024-look} benchmarks that explicitly consider the correctness and effectiveness of action executions.
\item Experimental results show that BacktrackAgent achieves improvements in task success rate and step accuracy on Mobile3M and Auto-UI benchmarks, and outperform the current SOTA methods MobileVLM \cite{wu-etal-2024-mobilevlm} and ReachAgent \cite{wu2025reachagentenhancingmobileagent}.
\end{itemize}

\section{Related Work}

\paragraph{GUI Agent.}
The rapid development of Large Language Models (LLMs) and Vision Language Models (VLMs) has created a strong foundation for developing GUI agents that can interact within digital environments \cite{liu2024autoglmautonomousfoundationagents,lin2024showuivisionlanguageactionmodelgui,gou2024navigatingdigitalworldhumans}. However, handling complex multi-step tasks remains a significant challenge \cite{liu2024agentbench,koh-etal-2024-visualwebarena,wang2025largeactionmodelsinception}. Many studies have explored various methods to improve the reasoning abilities of agents \cite{shen2024falconuiunderstandingguifollowing,putta2024agentqadvancedreasoning}. For example, EXACT \cite{yu2025exactteachingaiagents} and SWE-SEARCH \cite{antoniades2024swesearchenhancingsoftwareagents} utilize Monte Carlo Tree Search (MCTS) \cite{silver2016mastering} methods to enhance the decision-making processes. WebPilot \cite{zhang2025webpilot} generates a high-level plan for a task and continuously reflects on and refines that plan during the reasoning process. These methods heavily depend on the capabilities of VLMs like GPT-4o, neglecting whether the action executions align with the overall task goals. Mobile-Agent-E \cite{wang2025mobileagenteselfevolvingmobileassistant} introduces an Action Reflector to verify action outcomes and update the Tips and Shortcuts of the task. ReachAgent \cite{wu2025reachagentenhancingmobileagent} decomposes the task into subtasks and prioritizes the successful completion of these subtasks. InfiGUIAgent \cite{liu2025infiguiagentmultimodalgeneralistgui} reflects on whether the action results match expectations and generates a summary. Although these approaches utilize action execution outcomes as high-level guidance for the task, they still struggle with detecting and recovering from errors. In contrast, BacktrackAgent explicitly incorporates a backtracking mechanism to observe the outcomes of action executions, allowing it to detect and recover from error states effectively. 



\paragraph{Reinforcement Learning.}
Reinforcement learning (RL) techniques have been widely used to improve GUI agents \cite{chai2025a3androidagentarena}. DigiRL \cite{bai2024digirltraininginthewilddevicecontrol} and DistRL \cite{wang2024distrlasynchronousdistributedreinforcement} assign rewards to trajectory to help the agent align better with human preferences. ReachAgent \cite{wu2025reachagentenhancingmobileagent} samples step-level pairwise responses by utilizing Direct Preference Optimization (DPO). IPR \cite{xiong-etal-2024-watch} and UI-TARS \cite{qin2025uitarspioneeringautomatedgui} incorporate step-level supervision when training agents. 
BacktrackAgent directly uses the results from the error detection module as rewards to enhance the performance of the GUI agent. See Appendix \ref{appendix_related} for more related works.

\begin{figure*}[!t]

\setlength{\abovecaptionskip}{0.1cm}
\setlength{\belowcaptionskip}{-0.3cm}
\centering
\includegraphics[width=0.95\textwidth]{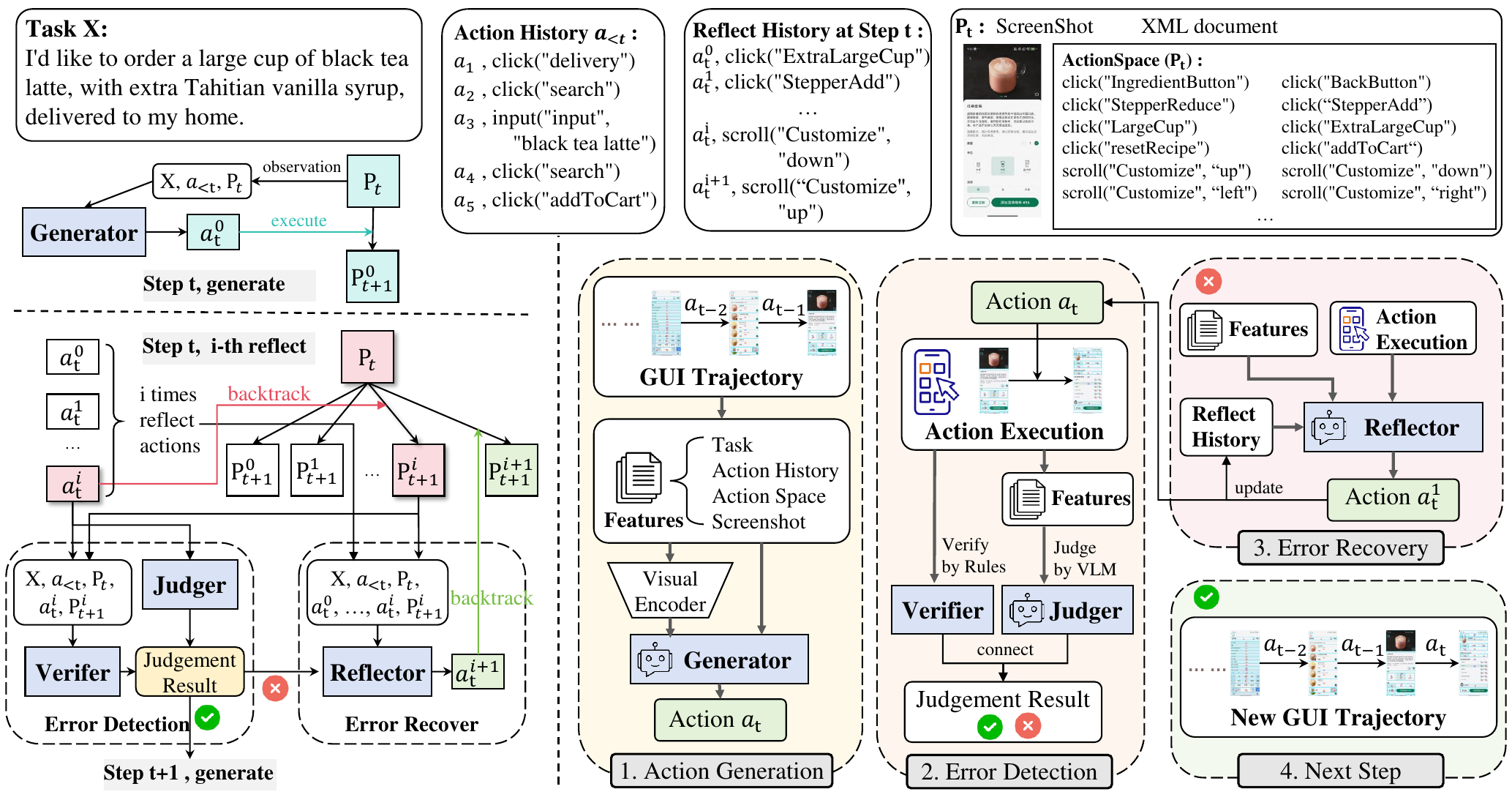}
\caption{The overview of BacktrackAgent. The left part shows the detailed process of an action $a^i_t$ being judged as an error by the error detection module and reflected by the error recovery module. The right part shows the pipeline of the agent generating GUI trajectories through action generation, error detection, and error recovery modules. }
\label{figure3-1}
\end{figure*}

\section{Methodology}\label{section3}

\subsection{Problem Formulation}


The goal is to simulate human behavior by performing multiple rounds of interactions with the GUI pages to complete a given task, referred to as task $\rm X$. Starting from the initial page $\rm P_1$, the agent observes the current GUI page $\rm P_t$ at each time step $t$ to generate an action $a_t$ that progresses towards completing the task. After executing $a_t$, the GUI environment updates, resulting in a new page $\rm P_{t+1}$. The sequence of executed actions is represented as $\textbf{a}=\{a_1, a_2,..., a_n\}$, while the sequence of corresponding GUI pages is represented as $\rm \textbf{P}=\{P_1, P_2,..., P_{n+1}\}$. The agent must ensure that the transitions between the GUI pages successfully lead to the completion of task $\rm X$.

\subsection{BacktrackAgent}

BacktrackAgent consists of four main modules: Generator, Verifier, Judger, and Reflector. 
Figure \ref{figure3-1} illustrates the inference process of BacktrackAgent at the $t$-th time-step:
\begin{enumerate}[itemsep=2pt,topsep=1pt,parsep=1pt]
\item In each interaction step, the Generator generates the current action $a_t$ based on task $\rm X$, the GUI page $\rm P_t$, and the history action list $\textbf{a}_{<t}=\{a_1,..., a_{t-1}\}$.
\item During error detection, a generated action $a^i_t$ is executed resulting in a new page $\rm P_{t+1}$, where $i$ represents the $i$-th reflection at time step $t$. The Verifier and the Judger assess whether $a^i_t$ is valid and contributes to completing task $\rm X$. Their evaluation considers the action $a^i_t$, the pages before and after execution ($\rm P_{t}$ and $\rm P_{t+1}$), as well as the relevant background information (task $\rm X$ and previous actions $\textbf{a}_{<t}$). The Verifier is a rule-based module that ensures that the action $a^i_t$ is executable and effective. The Judger is a model-based module that assesses whether executing action $a^i_t$ leads to an error page and if it improves the likelihood of achieving the task goal.
\item If both the Verifier and the Judger confirm that $a^i_t$ is correct, the agent considers it the final action at time-step $t$ and proceeds to time-step $t$+1. Otherwise, the agent goes to the Reflector for error recovery.
\item During error recovery, the Reflector updates the action $a^i_t$ to $a^{i+1}_t$ based on all reflected actions at time step $t$, as well as the pages before and after executing the action.
\end{enumerate}
BacktrackAgent repeats the above $2 \sim 4$ steps at each time step until $a^i_t$ is judged as correct or $i$ exceeds the max number of reflections. A step-by-step reasoning process refers to Appendix \ref{step_by_step_inference}.

\subsection{Modules}\label{section_modules}

\paragraph{Generator}



Given a task $X$, the generator generates action $a_t$ based on the GUI page $\rm P_t$, the extracted candidate action space, and the history actions $\textbf{a}_{<t}$ as follow:
\setlength\belowdisplayskip{4pt}
\begin{equation}\nonumber
\setlength\abovedisplayskip{4.5pt}
a_t = \text{Generator}(\rm X, P_t, \text{Acts}(P_t), \textbf{a}_{<t}),
\end{equation}
where $\rm \text{Acts}(P_t)$ denotes a list of all possible actions that can be performed on $\rm P_t$. Note that $a_t$ also belongs to $\rm \text{Acts}(P_t)$.





\paragraph{Verifier}
After the generator generates $a_t$, the agent simulates executing that action to update the page from $\rm P_t$ to $\rm P_{t+1}$. The verifier checks the effectiveness of $a_t$ based on two principles:

$\bullet$ The action must be valid and executable, falling into these categories: \{click, scroll, input, complete\}, and include properly formatted elements and parameters. Refer to Figure \ref{figure2-1} for examples.

$\bullet$ Upon executing an action, the environment should change as a result, except in cases where the task is complete. The verifier compares the pages $\rm P_t$ and $\rm P_{t+1}$; if they are identical, the action is considered ineffective.
\begin{equation}\nonumber
p^v_t = \text{Verifier}({\rm P_t, P_{t+1} }, a_t),
\end{equation}
where $p^v_t = 1$ indicates that the action is valid, and $p^v_t = 0$ indicates that it is not.

\begin{figure}[!t]
\setlength{\abovecaptionskip}{0.2cm}
\setlength{\belowcaptionskip}{-0.3cm}
\centering
\includegraphics[width=0.8\columnwidth]{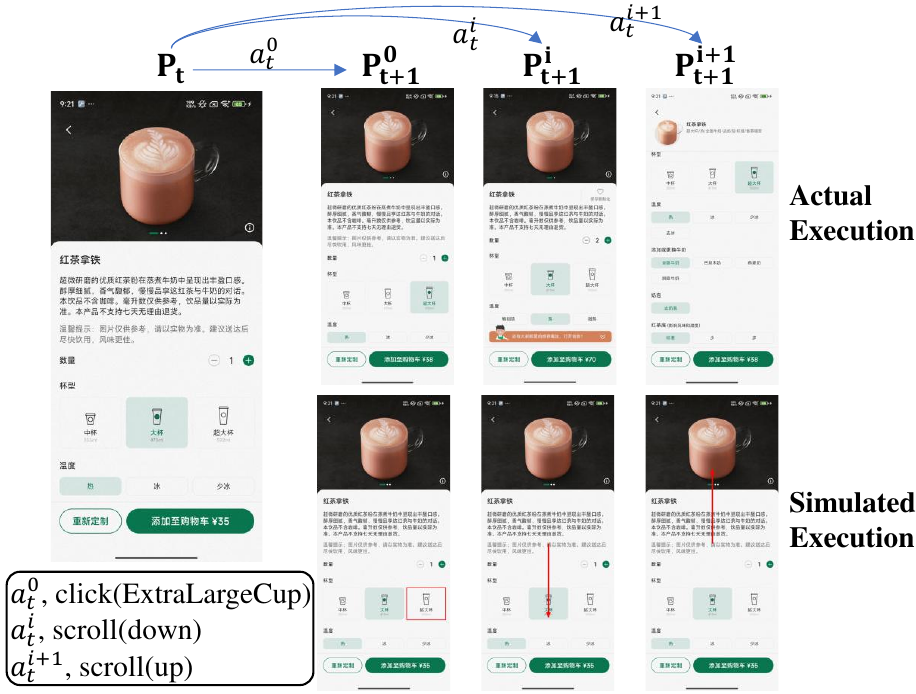}
\caption{The action result pages generated by actual execution and simulated execution.}
\label{figure4-1}
\end{figure}




\paragraph{Judger} 
With page $\rm P_t$, action $a_t$, and page $\rm P_{t+1}$, the judger assesses whether executing this action contributes to the successful completion of task $\rm X$. The judger functions as a binary classifier defined as follows:
\setlength\abovedisplayskip{4pt}
\begin{equation}\nonumber
p^j_t = \text{Judger}({\rm X, P_t, \text{Acts}(P_t), \textbf{a}_{<t}} , a_t, \rm P_{t+1}),
\end{equation}
where $p^j_t = 1$ indicates that the action is valuable, and $p^j_t = 0$ indicates that it is not. The prompt template for the judger is shown in Appendix \ref{appendix_A}.



\paragraph{Reflector}
The BacktrackAgent decides whether to modify the action based on the results from both the verifier and the judger. During the $i$-th rewriting process, if either the verifier or the judger determines that the action is ineffective or does not contribute to completing the task, the reflector generates a new action $a^i_t$ as follows:
\begin{equation}\nonumber
a^{i+1}_t \!\! = \! \text{Reflector}({\rm X, P_t, \text{Acts}(P_t), \textbf{a}_{<t}}, \textbf{a}^{\leq i}_t, \rm P^i_{t+1}  ),
\end{equation}
where $\textbf{a}^{<i}_t$ denotes all attempted actions $\textbf{a}^{\leq i}_t = \{a_t, a^1_t,..., a^{i}_t \}$ at time step $t$. The prompt template for the reflector is shown in Appendix \ref{appendix_A}. BacktrackAgent repeats the "verifier-judger-reflector" phase until both the verifier and the judger agree that the action is effective, or until the maximum number of rewrite iterations is reached.

\subsection{Action Execution} \label{execution_policy}


The process of performing the action $a_t$ and updating the GUI page from $\rm P_t$ to $\rm P_{t+1}$ is referred to as \textbf{Actual Execution}. However, for certain manually annotated datasets, we cannot reproduce the GUI environment and obtain page $\rm P_{t+1}$ that arises from executing a non-golden answer $a_t$. To address this issue, we identify the possible execution results of $a_t$ on page $\rm P_t$, such as drawing arrows for scroll actions and marking element boxes and input text for input actions. This process is called \textbf{Simulated Execution}. These annotated pages are then provided to the error detection and recovery modules to demonstrate the effects and potential impacts of the actions. As illustrated in Figure \ref{figure4-1}, after executing the click("ExtraLargeCup") action, the actual execution updates the cup type on the GUI page to an extra large cup. In contrast, the simulated execution marks the bounding box of the "ExtraLargeCup" element in red.

\subsection{Training}

We begin by using the multi-round page navigation task datasets to perform supervised fine-tuning of the VLM and to obtain the generator model. Next, we apply the generator model to the task datasets to create the training datasets for the judger and reflector models. All three models, the generator, judger, and reflector, are trained using cross-entropy loss as follows:
\setlength\abovedisplayskip{4pt}
\setlength\belowdisplayskip{4pt}
\begin{equation}\nonumber
\begin{split}
\mathcal{L}_g  \!= \!\! - \!\!\sum_{t}  \!\log \text{P}(&a_t |\rm X,  \!P_t,  \!\text{Acts}(P_t),  \!\textbf{a}_{<t}), \\
\mathcal{L}_j  \!= \!\! - \!\!\sum_{t}  \!\log \text{P}(&p^{j,i}_t | {\rm X,  \!P_t,  \!\text{Acts}(P_t),  \!\textbf{a}_{<t}} , a^i_t, \rm P^i_{t+1}), \\
\mathcal{L}_r  \!= \!\! - \!\!\sum_{t}  \!\log \text{P}  (&a^{i+1}_t | {\rm X,  \!P_t,  \!\text{Acts}(P_t),  \!\textbf{a}_{<t}}, \\
& \; \textbf{a}^{<i}_t, a^i_t, \rm P^i_{t+1}  ). \\ 
\end{split}
\end{equation}
Similar to value-based reinforcement learning methods such as DigiRL and DistRL, we score the actions generated by the generator and reflector at each step and use them to further reinforce the model. We directly use the results of the error detection module as the action rewards to feedback to the generator and reflector. The verifier loss and judger loss are defined as follows
\begin{equation}\nonumber
\mathcal{L}_{\text{verifier}} = 1 - p^v_t, \ \text{and} \ \mathcal{L}_{\text{judger}} = \text{P}(p^j_t = 0).
\end{equation}
The final loss $\mathcal{L}$ is a combination of the cross-entropy loss, the verifier loss, and the judger loss:
\begin{equation}\nonumber
\mathcal{L} = {\mathcal{L}}_{g} + \beta_1{\mathcal{L}}_{\text{verifier}} + \beta_2{\mathcal{L}}_{\text{judger}},
\end{equation}
where $\beta_1$ and $\beta_2$ are hyperparameters.

\section{Dataset Construction}

\subsection{Datasets}


We utilize the Mobile3M \cite{wu-etal-2024-mobilevlm, wu2025reachagentenhancingmobileagent} and Auto-UI \cite{zhang-zhang-2024-look} datasets. 
They are two largest public mobile control datasets, containing page navigation tasks that require multi-round interactions.
Mobile3M includes a total of 53,832 tasks with 259,725 action steps, while Auto-UI comprises 106,645 tasks and 988,518 action steps. 
Each task consists of a task instruction and a corresponding chained GUI trajectory, which includes a sequence of GUI pages and actions.

\begin{table}[!tb]
\setlength{\abovecaptionskip}{0.2cm}
\setlength{\belowcaptionskip}{-0.5cm}
\centering
\resizebox{0.9\columnwidth}{!}{
\begin{tabular}{lcc|cc}
\hline\hline
\multirow{2}{*}{Dataset} & \multicolumn{2}{c|}{Train} & \multicolumn{2}{c}{Test}   \\ \cline{2-3} \cline{4-5}
& Chain & Step & Chain & Step  \\  \hline
Mobile3M & 53,832 & 259,725 & 2,689 & 12,922  \\ \hline
Auto-UI & 106,645 & 988,518 & 55,780 & 450,924  \\ \hline\hline
\multirow{2}{*}{Dataset}	& \multicolumn{2}{c|}{Judger}		&	\multicolumn{2}{c}{Reflector}	\\ \cline{2-3} \cline{4-5}
	& Positive	& Negative		 & Positive & Negative \\ \hline 
Mobile3M		& 259,725	& 27,463	& 51,945 & 27,463	\\
Auto-UI		& 988,512	& 311,148	& 197,702 & 311,148	\\
\hline\hline
\end{tabular}}
\caption{The statistics of datasets.}
\label{table5_1}
\end{table}

\subsection{Datasets for Judger and Reflector}\label{dataset_judger_reflector}


To enhance the model's capability to detect and recover from error states, we utilize the training splits of Mobile3M and Auto-UI as seed datasets to construct SFT data for the Judger and Reflector. First, for each task in the training set, we employ the Generator to regenerate actions at the step level. Next, we simulate the execution of the generated actions on the current page to produce the subsequent page. We construct two datasets based on these actions and page information.

\noindent
\textbf{Judgment dataset. }
The judger's input consists of four parameters: task $\rm X$, the current GUI page $\rm P_t$, the current action $a_t$, and the subsequent GUI page $\rm P_{t+1}$. The output is a binary classification result indicating whether $a_t$ is effective in furthering the task completion on the current page. 
Since AutoUI is a chain-structured dataset and does not provide the complete XML document or images of the GUI environment, it is challenging to determine the resulting page after executing an incorrect action. We use simulated execution page as the subsequent page $\rm P_{t+1}$. Since Mobile3M is a graph-structured dataset and contains complete information on GUI pages, we use the actual execution page as $\rm P_{t+1}$. We construct a judgment dataset with incorrect actions generated by the generator and the golden answer from the original dataset. An effective generated action need to satisfy both the IoU and text metrics, as described in Section \ref{metric}.


\noindent
\textbf{Reflection dataset. }
The reflector must have two essential abilities: it should be able to correct any incorrect actions and preserve the correct actions that may be misjudged without making changes. After training the generator and judger using the original dataset and the judgment dataset, we utilize these two models to regenerate actions and judge their effectiveness. We then extract 100\% of the ineffective actions and 20\% of the effective actions to construct the reflection dataset.



The statistics of the original, judgment, and reflection datasets are summarized in Table \ref{table5_1}. The detailed judgment and reflection data construction process is shown in Appendix \ref{construct}.

\begin{table*}[!t]
    \centering

\setlength{\abovecaptionskip}{0.3cm}
\setlength{\belowcaptionskip}{-0.5cm}
\resizebox{1\textwidth}{!}{
    \begin{tabular}{ll|cc|c|ccc|cc}
    \hline\hline
        \multirow{3}{*}{Model}  & \multirow{3}{*}{Method}   &  \multicolumn{2}{c|}{Auto-UI}  & \multicolumn{6}{c}{Mobile3M }\\ \cline{3-10} 
        & & Task Level  & Step Level &   Task Success & \multicolumn{3}{c|}{Task Level Acc}  &  \multicolumn{2}{c}{Step Level Acc}   \\  \cline{6-10} 
        & & Accuracy & Accuracy &  Rate & Both &  IoU & Text & IoU  & Text   \\ \hline 
        GPT-4o & FewShot & 15.16  &	55.38   & - & - & - & - & 19.44  & 17.06   \\ 
        MobileVLM$\rm _{seperate}$   & FewShot &5.99  &	44.06 & - & - & - & - & 1.75  & 10.60   \\ \hline
        Qwen-VL & SFT & 16.97  &	68.75  & 35.77  & 20.58  & 30.13  & 26.22  & 73.38  & 72.14   \\ 
        Auto-UI$\rm _{unified}$   & SFT &  24.79 &	75.13 & 33.40  & 18.40  & 29.60  & 22.20  & 73.26  & 70.88   \\ 
        MobileVLM & SFT & 25.53 &	77.36   & 39.78  & 22.68  & 34.03  & 28.43  & 76.20  & 74.08   \\
        Qwen2-VL  & SFT &21.56 &	72.26  & 44.81  & 27.48  & 34.88  & 30.87  & 81.87  & 80.64   \\  
        ReachAgent$\rm _{stage 1}$  & SFT & 24.89 &	74.54 &  45.33  & 27.48  & 37.82  & 31.31  & 83.34  & 81.47   \\ 
        ReachAgent$\rm _{stage 2}$   & SFT+RL & 25.28 &	74.81  & 46.52  & 29.79  & 38.75  & 33.06  & 83.32  & 81.77   \\ \hline
        BacktrackAgent & SFT+RL & \textbf{29.72 } &	\textbf{78.04 } & \textbf{54.11 } & \textbf{33.51 } & \textbf{43.25 } & \textbf{36.67 } & \textbf{84.94 } & \textbf{83.24 } \\ \hline\hline
    \end{tabular}}
    \caption{Main Result(\%) on Auto-UI and Mobile3M benchamrks. - denotes less than 1\%.
    }
    \label{result_1}
\end{table*}

\section{Experiment}

\subsection{Benchmarks and Metrics} \label{metric}

We use the official test sets of Mobile3M and Auto-UI to evaluate BacktrackAgent for comparison. There is no overlap between the training and testing datasets.  We use three metrics for evaluation.

$\bullet$ Task Success Rate evaluates GUI trajectories at the task level. When a GUI trajectory contains the final page in the golden answer, it is considered as navigating to the key page and successfully completing the task.

$\bullet$ Task Level Accuracy evaluates whether each GUI trajectory is consistent with the golden trajectory. Only when all actions in the GUI trajectory match the golden answer is it considered a task-level match. IoU and Text metrics use bounding box parameters and text parameters to compare the generated actions with the golden actions.

$\bullet$ Step Level Accuracy evaluates whether each generated action is consistent with the golden action at the step level.



\subsection{Parameters and Baselines}


BacktrackAgent uses Qwen2-VL-7B as the backbone model. The Generator, Judger, and Reflector were trained for 2 epochs in the SFT version. The Generator and Reflector were further trained for 2 epochs in the RL version. To ensure a fair comparison, all baselines and variants of BacktrackAgent maintain consistent hyperparameters.

We compare our approach with the following strong baselines: GPT-4o, Auto-UI, Qwen-VL, Qwen2-VL, MobileVLM, and ReachAgent. Except for GPT-4o, all baselines use the backbone model with 7B parameters. For more detailed information, refer to Appendix \ref{appendix_setting} and \ref{appendix_baselines}.

\begin{table*}[htbp]\small
    \centering

\setlength{\abovecaptionskip}{0.3cm}
\setlength{\belowcaptionskip}{-0.5cm}
\resizebox{0.9\textwidth}{!}{
    \begin{tabular}{l|c|ccc|cc}
    \hline\hline
        \multirow{2}{*}{Model}   &  Task Success & \multicolumn{3}{c|}{Task Level Acc}  &  \multicolumn{2}{c}{Step Level Acc}   \\  \cline{3-7} 
         &  Rate & Both &  IoU & Text & IoU  & Text   \\  \hline\hline
\multicolumn{7}{c}{\textbf{Backtrack Mechanism}} \\  \hline
BacktrackAgent w/o Judger \& Verifier \& Reflector	 	 &48.46	 &29.56	 &38.90	 &33.06	 &83.22	 &81.72\\  
BacktrackAgent w/o Judger	 	 &48.79	 &29.90	 &39.20	 &33.32	 &83.35	 &81.84\\  
BacktrackAgent w/o Verifier	 	 &53.66	 &32.99	 &42.73	 &36.30	 &84.75	 &83.12\\  \hline
\textbf{BacktrackAgent}	  &\textbf{54.11} & \textbf{33.51} & \textbf{43.25} & \textbf{36.67} & \textbf{84.94} & \textbf{83.24} \\ 
\;\;$\Delta$ Backtrack Mechanism	 	 &5.65	 &3.95	 &4.35	 &3.61	 &1.72	 &1.52\\  
\;\;$\Delta$ Judger	 	 &5.32	 &3.61	 &4.05	 &3.35	 &1.59	 &1.40\\  
\;\;$\Delta$ Verifier	 	 &0.45	 &0.52	 &0.52	 &0.37	 &0.19	 &0.12\\    \hline\hline
\multicolumn{7}{c}{\textbf{RL Mechanism}} \\  \hline
BacktrackAgent w/o RL	 	 &52.18	 &32.58	 &41.61	 &35.48	 &84.67	 &82.84\\  \hline
\textbf{BacktrackAgent}	  &\textbf{54.11} & \textbf{33.51} & \textbf{43.25} & \textbf{36.67} & \textbf{84.94} & \textbf{83.24} \\ 
\;\;$\Delta$ RL	 	 &1.93	 &0.93	 &1.64	 &1.19	 &0.27	 &0.40\\    \hline\hline
\multicolumn{7}{c}{\textbf{Execution Method}}  \\   \hline
BacktrackAgent w/o Backtrack	 	 &48.46	 &29.56	 &38.90	 &33.06	 &83.22	 &81.72\\  
BacktrackAgent-Simulate Execution 	 &49.16	 &28.93	 &39.01	 &32.54	 &83.16	 &81.43\\   \hline
\textbf{BacktrackAgent}	  &\textbf{54.11} & \textbf{33.51} & \textbf{43.25} & \textbf{36.67} & \textbf{84.94} & \textbf{83.24} \\ 
\;\;$\Delta$ Backtrack with Simulate Execution 	 \;\;\;\;\;\;\;\;\;	 &0.70	 &-0.63	 &0.11	 &-0.52	 &-0.06	 &-0.29\\  
\;\;$\Delta$ Backtrack with Acutal Execution	 	 &5.65	 &3.95	 &4.35	 &3.61	 &1.72	 &1.52 \\ \hline\hline
    \end{tabular}}
    \caption{
    Ablation study (\%) on the Backtrack, the RL Mechanism, and execution policy.
    Here, Simulate/Actual means that the Judger and Reflector obtain the next page $P^i_{t+1}$ by actually/simulating execution action $a^i_t$. 
    }
    \label{result_ablation}
\end{table*}

\subsection{Main results}
The main experimental results are shown in Tables \ref{result_1}. We can observe that:

$\bullet$ For the Mobile3M benchmark, at the task level, BacktrackAgent improves the task success rate by 7.59\% and the task level accuracy by 3.72\%. We attribute this to the backtracking mechanism with the judge, verifier, and reflector. BacktrackAgent achieves better performance by learning to detect and recover from erroneous pages.

$\bullet$ Compared with the DPO-based ReachAgent, BacktrackAgent improves the step-level IoU accuracy and text accuracy by 1.64\% and 1.47\%, respectively. This proves that the explicit backtracking can better capture the agent's errors and further improve the agent's performance compared to the pre-sampled paired positive and negative data.

$\bullet$ For the Auto-UI benchmark, BacktrackAgent outperforms the SOTA baseline in both step-level and task-level. The improvement of BacktrackAgent proves that our framework and backtracking mechanism can generally improve task completion abilities on different datasets. For more detailed results on Auto-UI, please refer to Appendix \ref{appendix_autoui}.

\subsection{Ablation Study}
\label{ablation}

To better evaluate the effect of each module, we conducted several ablation experiments, as shown in Table  \ref{result_ablation}.
From the table, we can see that:

$\bullet$ The backtracking mechanism improves the task success rate by 5.65\% and the accuracy at both task-level and step-level by more than 3.5\% and 1.5\%, respectively. This is because backtracking helps the agent better align the action execution results with the task goals, enabling the agent to detect and correct errors.

$\bullet$ Compared with the verifier, the judger contributes more to performance improvement. This is because while the verifier can accurately detect invalid actions and correct these unnecessary errors, as the agent's performance improves, the probability of generating invalid actions decreases, resulting in a relatively small overall improvement.

$\bullet$ The reinforcement learning improves the performance in all indicators, especially in task success rate and task-level accuracy, which are improved by 1.93\% and 0.93\% respectively. The two additional losses help the agent better align with the preferences of the verifier and judger, thereby improving its ability to complete tasks.

$\bullet$ The backtracking mechanism trained with the actual execution page outperformed the one trained with the simulated execution page. Compared with the 5.65\% increase in task success rate caused by the actual page, the simulated page only achieved a 0.7\% increase in this metric and caused a decrease in task-level and step-level accuracy. The reason is that the actual execution page provides a more accurate representation of the execution results, which enables the error detection module to more effectively identify deviations from the task goal.

\subsection{Stability and Applicability Analysis}

\begin{figure}[!t]
\setlength{\abovecaptionskip}{-0.2cm}
\setlength{\belowcaptionskip}{-0.3cm}
\centering
\includegraphics[width=0.9\columnwidth]{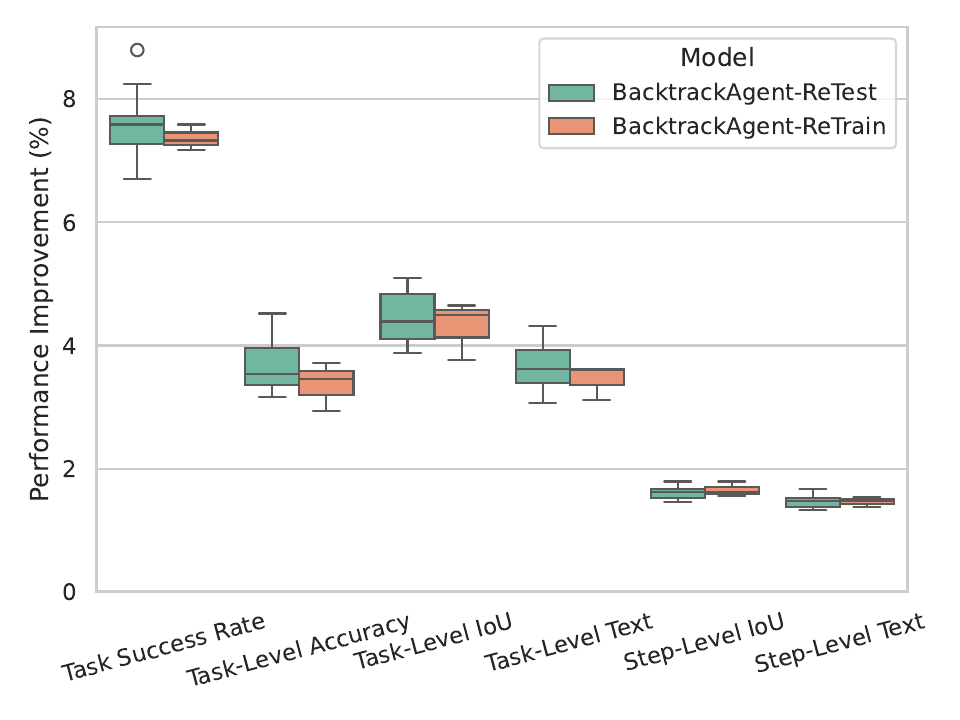}
\caption{Box plot shows the performance improvement (\%) of repeated experiments compared to ReachAgent.} 
\label{figure-boxplot}
\end{figure}

\begin{table}[!t]\small
\setlength{\abovecaptionskip}{0.3cm}
\setlength{\belowcaptionskip}{-0.3cm}
    \centering
\resizebox{1\columnwidth}{!}{
    \begin{tabular}{l|c|c|c}
    \hline\hline
        Dataset &  \multicolumn{2}{c|}{Mobile3M} & Auto-UI  \\   \hline
        Execution & Simulated & Actual & Simulated  \\  \hline
        Speed Ratio & 0.451x & 0.517x & 0.482x \\
        Total & 2.172 & 1.938 & 1.374  \\   \hline
        Generator & 0.979 & 1.002 & 0.663  \\ 
        Judger & 0.803 & 0.806 & 0.296  \\ 
        Verifier & 0.003 & 0.003 & 0.004  \\ 
        Reflector & 0.129 & 0.127 & 0.179  \\ 
        Action Execution & 0.258 & - & 0.232 \\ \hline\hline
    \end{tabular}}
    \caption{Effiency(s/step) of BacktrackAgent. The speed ratio is the ratio of the time required for a step with the entire agent to the time required with just the Generator.
    }
    \label{result_4}
\end{table}

\textbf{Stability.}
To explore whether BacktrackAgent can stably maintain its performance advantage under different parameters, we conducted repeated experiments. For the training phase, we retrained the agent twice  from the backbone model with different seeds. For the testing phase, we repeated the test 10 times, randomly sampling 80\% of the test samples each time.
The box plots in Figure \ref{figure-boxplot} show the performance improvement of these experiments compared to the SOTA ReachAgent. BacktrackAgent's task success rate is 7.59\% higher than ReachAgent, and the performance fluctuation caused by the resampled test set and retrained Agent is less than 1.2\%. Since the step-level accuracy is already over 80\%, there is limited room for further improvement. However, BacktrackAgent's step accuracy has still improved by 1.62\% and 1.47\%, and the fluctuation in repeated experiments is less than 0.2\%. This shows that our BacktrackAgent is reliable and stable. More experimental data can be found in Appendix \ref{appendix_significant}.


\textbf{Time Efficiency.} 
Table \ref{result_4} presents the average time taken by each module of the BacktrackAgent during inference. The inference efficiency of the agent utilizing the backtracking mechanism is approximately 50\% of that of other agents that rely solely on a generator. The judger requires more time than the reflector because only the wrong actions need to be rewritten. Simulated action execution takes about $0.25$ seconds, as a new screenshot needs to be saved, while the speed of real action execution depends on the GUI environment itself. Overall, although the backtracking mechanism reduces inference speed, it remains valuable due to its significant contribution to the agent's ability to complete tasks effectively.

\begin{table}[!t]\small
\setlength{\abovecaptionskip}{0.2cm}
\setlength{\belowcaptionskip}{-0.3cm}
    \centering
    \begin{tabular}{l|ccc}
    \hline \hline
        \multicolumn{4}{c }{Error Detection}    \\ \hline
        ~ &  Precision & Recall & F1  \\ \hline
        Mobile3M & 75.12\% & 43.58\% & 55.16\%  \\ 
        Auto-UI & 80.01\% & 48.04\% & 60.04\%  \\ \hline  \hline
        \multicolumn{4}{c }{Error Recovery}  \\ \hline
        ~ &   Both & IoU & OCR  \\ \hline
        Mobile3M & 38.93\% & 49.90\% & 43.39\%  \\  
        Auto-UI & 31.24\% & 31.43\% & 31.61\% \\ \hline\hline
    \end{tabular}
    \caption{Accuracy (\%) of error detection and recovery modules of BacktrackAgent.
    }
    \label{result_5}
\end{table}

\textbf{Performance of the Error Detection and Recovery Modules.} 
Figure \ref{result_5} shows the accuracy of BacktrackAgent in detecting and recovering from errors.  Recall measures how many wrong actions are successfully detected by the agent, and Precision measures whether the actions detected as wrong are indeed wrong.  On Mobile3M, BacktrackAgent can detect 43.58\% of error actions and guarantee the accuracy of 75.12\% of all detected errors. The error recovery module can correct 38.93\% of these detected actions. On Auto-UI, BacktrackAgent achieves better error detection performance but worse error recovery performance. Refer to Appendix \ref{case} for case study.

\begin{table}[!t]\small
\setlength{\abovecaptionskip}{0.2cm}
\setlength{\belowcaptionskip}{-0.4cm}
    \centering
\resizebox{0.9\columnwidth}{!}{
    \begin{tabular}{l|cc}
    \hline \hline
        \multicolumn{3}{c }{Error Detection}    \\ \hline
        ~ &  Actually Error &  Actually Correct  \\ \hline
        Judge as Error & \textbf{8.48\%} & \textbf{2.81\%}   \\  
        Judge as Correct & 10.98\% & 77.73\%  \\ \hline  \hline
        \multicolumn{3}{c }{Error Recovery of “Judge as Error” Data}  \\ \hline
        ~ &   Actually Error &  Actually Correct  \\ \hline
        Correctly Recover & \textbf{2.37\%} & 2.03\%  \\  
        Failed to Recover & 6.11\% & 0.78\% \\ \hline\hline
    \end{tabular}}
    \caption{Distribution of error detect and recover modules on Mobile3M dataset.
    }
    \label{result_6}
\end{table}

Table \ref{result_6} analyzes the distribution of all generated results of Mobile3M after error detection and recovery.
We can see that the error detection module judged 11.29\% of the generated results as errors. Among them, 8.48\% of the actions were indeed wrong, and 2.81\% of the actions were misjudged by the error detection module.
For the 8.48\% of wrong actions the model successfully recovered 2.37\%, leaving 6.11\% unrecovered. For the 2.81\% of misjudged actions, the error recovery module incorrectly modified 0.78\%.
Overall, the performance of the BacktrackAgent is improved through error detection and recovery mechanisms.

\begin{table}[!t]
\setlength{\abovecaptionskip}{0.3cm}
\setlength{\belowcaptionskip}{-0.4cm}
    \centering
\resizebox{1\columnwidth}{!}{
    \begin{tabular}{l|cc|cc|cc|c}
    \hline \hline
       \multirow{2}{*}{Accuracy} & \multicolumn{2}{c|}{Click}  & \multicolumn{2}{c|}{Scroll} &  \multicolumn{2}{c|}{ Input}   & Complete \\ \cline{2-8} 
& IoU & Text & IoU & Text & IoU & Text &   \\ \hline
Percentage & \multicolumn{2}{c|}{79.24\%}  & \multicolumn{2}{c|}{15.10\%}  & \multicolumn{2}{c|}{4.84\%}  & 26.06\%  \\  \hline
ReachAgent & 82.09 & 83.12 & 71.25 & 55.07 & 92.80 & 88.80 & 91.82   \\
BacktrackAgent & 83.52 & 84.46 & 72.40 & 55.33 & 91.20 & 86.80 & 95.02   \\
$\Delta$   & 1.43 & 1.34 & 1.15 & 0.26 & -1.60 & -2.00 & 3.20    \\ \hline\hline
    \end{tabular}}
    \caption{ Statistical results of different types of actions.
    }
    \label{result_7}
\end{table}

\textbf{Action Types Analyze.}
From Table \ref{result_7}, we can see that:
1. The scroll action is most likely to be generated incorrectly. Even if the agent successfully selects the scroll action, it is difficult to generate the direction correctly. This is because the agent generating a scroll action usually means there are no available elements in the current page and needs to explore other pages, and this exploration action may not be unique.
2. Compared with ReachAgent, BacktrackAgent improves accuracy in click, scroll, and complete actions, but decreases in input actions. This is because the page changes after the input action are not obvious. In addition, the keywords of the input action are more likely to be changed to words that appear in the task when backtracking. However, the probability of input actions in GUI tasks is low (4.84\%), so the overall performance of the agent is still improved.

\section{Conclusion}



In this paper, we introduce BacktrackAgent, a framework that utilizes a backtracking mechanism to enhance the task completion capabilities of GUI agents. Our framework incorporates two error detection modules: verifier and judger, along with a recovery module: reflector, which explicitly handles the backtracking process following an erroneous action. Additionally, the rewards from the verifier and judger are integrated to further improve BacktrackAgent's performance. The experimental results show that BacktrackAgent increases the task success rate by 7.59\%. It also enhances the accuracy at both the task and step levels by 3.72\% and 1.64\%, respectively. By explicitly incorporating the backtracking mechanism, BacktrackAgent demonstrates superior performance in task completion. We hope that this agent framework will serve as a valuable resource for error detection and recovery tasks, contributing to future research in the community.

\clearpage

\section*{Limitations}


Despite the great progress made by BacktrackAgent, it still has some limitations that may be addressed in future updates. When performing the GUI tasks, our framework requires extra error detection and recovery modules, which reduces the agent's reasoning speed by 50\%. However, the substantial contribution of the backtracking mechanism to task completion gives us confidence in its potential for future improvements.

\section*{Ethics Statement}


This paper is conducted in accordance with the ACM Code of Ethics. The Mobile3M and Auto-UI datasets utilized in this research are publicly available. Our dataset for judger and reflector has been constructed using publicly available platforms and data sources, which ensures that there are no privacy issues or violations. All data used in our research is obtained following legal and ethical standards, and we do not collect any personally identifiable information. We will open-source all training and test data once the paper is accepted.

\bibliography{acl_latex}

\newpage
\appendix

\section{Experiment Settings} 
\subsection{Datasets}  \label{appendix_datasets}

Mobile3M is a pre-trained dataset collected on 49 third-party real-world apps using a breadth-first exploration method. 
Mobile3M collects data in a random exploration manner and constructs the GUI Trajectory of each APP in the form of a graph. This allows us to obtain various possible next pages for every GUI Page, depending on the actions taken, resulting in diverse GUI trajectories. ReachAgent filters GUI trajectories and annotation tasks from Mobile3M and reconstructs them into a page navigation dataset. Mobile3M traverses and executes each action in the action space of each GUI page when it is built, and marks the equivalent pages. Therefore, we can get the actual execution results from Mobile3M. If a generated action is not in the action space, we regard it as an invalid action.

Auto-UI cleans and extracts data from AITW dataset \cite{rawles2023android}, including 5 different types of tasks, General, GoogleApps, Install, Single, and WebShopping. These five types of tasks are quite different, so the agent trained on the five subsets separately performs better than the unified model trained on all five subsets, as shown in Table \ref{result_1}. Here, since the Auto-UI dataset does not contain the complete XML document of the GUI page or the mobile environment image, it is difficult for us to obtain the result page after executing a wrong action on the GUI page, so we use the simulated execution page as the result of the action execution.

\begin{table}[!htb]
    \centering
\resizebox{\columnwidth}{!}{
    \begin{tabular}{l|c|c}
        \hline
Hyperparameter & SFT  & RL \\
        \hline
epoch & 2 & 2 \\
batch size & 2 & 1 \\
learning rate & 1e-5 & 1e-5\\
warmup ratio & 0.1 & 0.1\\
max sequence length & 8192 & 8192 \\
max new tokens & 512 & 512 \\
GPUs  & 8 & 8 \\
num workers  & 128 & 128 \\
optimizer & Adam & Adam \\
deepspeed & ZeRO3 & ZeRO2 \\
max reflection times & 3 & 3 \\
$\beta_1$ & - & 0.1 \\
$\beta_2$ & - & 0.1 \\
         \hline
    \end{tabular}}
    \caption{Hyperparameters.}
    \label{parameters}
\end{table}

\begin{tcolorbox}[title = {3 Examples of GUI Trajectory Pairs}]

Task1: Search for today's gold price.

****Generate GUI Trajectory:****

Click(box1, "Search Box")

Input(box2, "Gold Price")

Click(box3, "Search Button")

****Golden GUI Trajectory:****

Click(box1, "Search Box")

Input(box2, "Today's Gold Price")

Click(box3, "Search Button")

---------------------------------------------------

Task2: Set the display mode to night mode.

****Generate GUI Trajectory:****

Click(box1, "Personal Center")

Click(box2, "Setting")

Click(box3, "Display Mode")

Click(box4, "Night Mode")

****Golden GUI Trajectory:****

Click(box1, "Personal Center")

Click(box3, "Display Mode")

Click(box4, "Night Mode")

---------------------------------------------------

Task3: Add Black Tea Latte to cart.

****Generate GUI Trajectory:****

Click(box1, "Search Box")

Input(box2, "Black tea Latte")

Click(box3, "Black tea Latte")

Click(box5, "Add to Cart")

****Golden GUI Trajectory:****

Click(box1, "Search Box")

Input(box2, "Latte")

Click(box4, "Black tea Latte")

Click(box5, "Add to Cart")

\end{tcolorbox}

\subsection{Metrics} \label{appendix_metric}

We evaluate the model performance at two levels: step level and task level. At the step level, we evaluate whether the generated action is correct in each time step. At the task level, we assess whether a GUI trajectory meets the requirements of the task. 

$\bullet$  Step Level Accuracy: Following ReachAgent, we use IoU accuracy to evaluate the intersection ratio between the bounding boxes in the generated and golden actions, allowing a 14\% error. Text accuracy evaluates whether the text in the generated action is consistent with that in the golden action, requiring F1 to be greater than $0.8$.

$\bullet$ Task Level Accuracy:  Task accuracy requires that each action in the GUI trajectory exactly matches the predetermined correct sequence. 

$\bullet$ Task Success Rate: Task Success Rate indicates whether the GUI trajectory navigates through the essential pages and completes the specified operations of the task. Following ReachAgent, if the GUI reaches the key page via a different route or continues to navigate after completing the task, we still consider the task to be successfully completed.

The table above shows 3 examples of GUI trajectory pairs. In Task 1, the second step shares the same bounding box but different text, so this step matches on the IoU metric but not on the Text metric. In Task 2, the agent's actions from the second step onwards are not completely consistent with the ground truth, so only one of the three steps is a step-level match. In Task 3, the second step matches on the IoU metric but not on the Text metric, and the third step matches on the Text metric but not on the IoU metric.

Therefore, their step-level metrics can be calculated as follows: 

$\bullet$  Step Level Accuracy-IoU: (3+1+3)/(3+3+4)

$\bullet$  Step Level Accuracy-OCR: (2+1+3)/(3+3+4)

In addition, all three tasks successfully reached the final page of the golden answer. However, Task 1 is completely consistent with the golden answer only on IoU metric. The remaining two tasks are not completely consistent with the golden answer on both IoU and Text metrics. Therefore, their task-level metrics can be calculated as follows:

$\bullet$ Task Success Rate: 3/3

$\bullet$ Task Level Accuracy-Both: 0/3

$\bullet$ Task Level Accuracy-IoU: 1/3

$\bullet$ Task Level Accuracy-OCR: 0/3

\subsection{Parameters}\label{appendix_setting}
The hyperparameters are presented in Table \ref{parameters}.
BacktrackAgent uses Qwen2-VL-7B as the backbone model. We use 8 80GB Nvidia A100 GPUs for fine-tuning. Here, 2 epochs of fine-tuning typically cost 25 hours on Mobile3M and 97 hours on Auto-UI. The learning rate is 1e-5. The agent's max length is 8192. $\beta_1$ and $\beta_2$ is 0.1. The maximum number of reflections for each step is 3.  For the SFT version, the Generator, Judge, and Reflector were trained for 2 epochs on the Mobile3M and Auto-UI datasets, respectively. For the RL version, the generator and reflector were further trained for 2 epochs with the new loss function. To ensure fair comparisons, we maintain consistent hyperparameters across all the baselines and the ablations of BacktrackAgent. 

For the Mobile3M dataset, the generator first trained for 2 epochs on 259,725 data with a batch size of 2. The judger and the reflector were trained for 2 epochs in the constructed dataset as described in Section \ref{dataset_judger_reflector}. Then, the generator and the reflector were further finetuned for 2 epochs with additional loss from the error detection module with a batch size of 1. During testing, the max reflection time is set to 3. 

For the Auto-UI dataset, We fine-tune BacktrackAgent on 5 subsets respectively. Similarly, we fine-tuned the generator, judger, and reflector for 2 epochs. Then, we further reinforced the generator and reflector for 2 epochs.

\subsection{Baselines} \label{appendix_baselines}
We compare our proposed BacktrackAgent with the following baselines: GPT-4o, Auto-UI, Qwen-VL, Qwen2-VL, MobileVLM, and ReachAgent.

\begin{itemize}
    \item GPT-4o \cite{openai2023gpt4} is a large available VLM and has been widely used in the development of agents \cite{yu2025exactteachingaiagents,zhang2025webpilot}.
    \item Qwen-VL \cite{bai2023qwen} is a large-scale vision-language model with open weights. It is used as the backbone model for multiple mobile AI agents.
    \item Qwen2-VL \cite{Qwen2-VL} is an improved version of Qwen-VL. It can understand images of different resolutions and has the ability of complex reasoning and decision-making.
    \item Auto-UI \cite{zhan2023you} is a GUI agent that focuses on action history and future action plans
    \item MobileVLM \cite{wu-etal-2024-mobilevlm} uses a large number of randomly explored pages from Mobile3M for two-stage pre-training, which improves its ability to understand the elements within a page and the relationships between pages.
    \item ReachAgent \cite{wu2025reachagentenhancingmobileagent} is a GUI agent that focuses on page reach and page operation subtasks. It further enhances the model's task completion abilities by building pairwise responses based on the DPO method.
\end{itemize}
For in-context learning like GPT-4o, we provided them with several few-shot examples. For other baselines, we use the same training dataset to supervise fine-tune them for two epochs.

\subsection{Verifier's Rules} \label{appendix_verifier}

As described in Section \ref{section_modules}, we have formulated two very general rules in Verifier that should be applicable to a variety of different GUI environments.

For Rule 1, we require that the action be complete and executable. Regardless of the GUI platform (Mobile, Desktop, Web) and the format in which the action is organized (Action, API, Code), an executable action should be the foundation of a valid GUI interaction. 

For Rule 2, we require that the page will change after the action is executed. This rule ensures that the operation can truly affect the GUI environment and is generally applicable across different GUI environments.

Considering the generality of these two rules, we believe they can be extended to various GUI environments.



\begin{table*}[!t]
\centering
\resizebox{\textwidth}{!}{
\begin{tabular}{l}
\toprule
\colorbox{gray!20}{\{image\}}  \\
The actions you can use are: \\
\colorbox{gray!20}{\{action space\}} \\
You need to complete the following task: \\
\colorbox{gray!20}{\{task\}} \\
The completed actions are as follows: \\
\colorbox{gray!20}{\{history actions\}} \\
Judgment: Please analyze whether the next action is helpful to further complete the task based on the \\ current status and completed actions. \\
Next action: \colorbox{gray!20}{\{next action\} }\\

The page changes caused by executing the action are as follows: \\
\colorbox{gray!20}{\{image\}}  \\

Final judgment (whether the next action is helpful to complete the task): (Yes or No)\\

\toprule
\end{tabular}
}
\caption{The prompt for the judger in the BacktrackAgent. The grey text indicates the page information and history actions to be filled in.}
\label{prompt_judgment}
\end{table*}

\begin{table*}[!t]
\centering
\resizebox{\textwidth}{!}{
\begin{tabular}{l}
\toprule
\colorbox{gray!20}{\{image\}}  \\
The actions you can use are: \\
\colorbox{gray!20}{\{action space\}} \\
You need to complete the following task: \\
\colorbox{gray!20}{\{task\}} \\
The completed actions are as follows: \\
\colorbox{gray!20}{\{history actions\}} \\
Reflection: This is not your first attempt to generate the next action. 
The previous attempts to generate \\ the next action have all failed. Here are some previously generated next actions: \\

\colorbox{gray!20}{\{next actions\} }\\

The page changes caused by executing the action are as follows: \\
\colorbox{gray!20}{\{image\}}  \\

Please note that you are currently in the middle stage of the trajectory. First, you need to analyze the \\ current state, completed actions, and tasks, and compare them 
 with the previous attempts at the next \\ action. Then, you need to generate a new action that is different from all previously generated next \\ actions.\\
\toprule
\end{tabular}
}
\caption{The prompt for the reflector in the BacktrackAgent.  }
\label{prompt_reflection}
\end{table*}

\begin{table}[!t]
\setlength{\belowcaptionskip}{-0.5cm}
\centering
\begin{tabular}{l}
\toprule
\colorbox{gray!20}{\{image\}}  \\
The actions you can use are: \\
\colorbox{gray!20}{\{action space\}} \\
You need to complete the following task: \\
\colorbox{gray!20}{\{task\}} \\
The completed actions are as follows: \\
\colorbox{gray!20}{\{history actions\}} \\
\toprule
\end{tabular}
\caption{The prompt for the agent's basic generator. The grey text indicates the page information and history actions to be filled in.}
\label{prompt_action}
\end{table}

\section{Releated Work}\label{appendix_related}

In this section, we will discuss other reflection/verifier/backtracking mechanisms used in LLM-agents and their similarities and differences with BacktrackAgent.

\paragraph{Reflection.}

Some past works have adopted reflection for self-improvement, improving generation through self-evaluation during reasoning \cite{madaan2023self}.
Reflection \cite{shinn2023reflexion} leverages verbal reinforcement to teach agents to learn from past mistakes. Specifically, after performing an action, it observes the state of the current environment, generates feedback in the form of a text summary, and provides it to the agent as additional context when generating the next action. Similarly, Mobile-Agent-v2 \cite{wang2024mobile} adopts a reflection agent to observe the screen state before and after the decision agent's operation to determine whether the current operation is effective, so as to avoid falling into a loop of invalid operations. Mobile-Agent-E \cite{wang2025mobileagenteselfevolvingmobileassistant} generate plans and shortcuts for GUI tasks and continuously reflect and update these hints during reasoning. These methods use VLMs such as GPT-4o and GPT-4V as core models, and formulate different prompts to encourage the model to analyze the results of previous actions.  They rely heavily on the ability and performance of the core model, the quality of the prompts, and they also have difficulty in encouraging action execution to be consistent with the overall task goal.

\paragraph{Verifier.}

Previous work has demonstrated the effectiveness of verifiers in the fields of math question answering and code generation.
For math question answering tasks \cite{cobbe2021training, shen2021generate}, models can execute mathematical expressions to avoid generating malformed results or using variables not mentioned in the question.
For code generation tasks \cite{li2022competition, chen2022codet, chen2023teaching}, models simulate the execution of generated code with test cases or self-generated unit tests to detect and fix errors in the program.
Some work uses methods such as reinforcement learning or scoring models \cite{le2022coderl, ni2023lever} to further improve existing generation based on feedback or scoring of execution results.

A main reason why verifiers are effective on these tasks is that both mathematical expressions and codes are executable, and the results after execution reflect the quality of generation.
Similarly, in GUI scenarios, the interactive actions generated by the agent at each step are executable and the results of the execution can be observed. The changes in the GUI page can also reflect whether the generated actions are relevant to the task and effective.

The setting of BacktrackAgent is closer to LEVER [10], which trains a judger separately and judges the results before and after the model execution to guide the backtrack mechanism. The judger and the rule-based verifier jointly judge whether the generated actions are consistent with expectations and helpful for task completion.

\paragraph{Backtrack Mechanism.}
Some works further utilize backtracking algorithms to explicitly intervene in the reasoning process. Mobile-Agent-v2 \cite{wang2024mobile} detects whether the current action is wrong or invalid and regenerates these incorrect actions. Neither wrong nor invalid actions are recorded in the action history to prevent the agent from tracking these operations. WebPliot \cite{zhang2025webpilot} uses an MCTS-based approach to explore the action space of Web tasks. It uses the maximum backpropagation (MVB) mechanism to prioritize the most promising paths for the MCTS backpropagation step. 

Our BacktrackAgent adopts a rule-based verifier and a model-based judger to jointly guide the backtracking mechanism. It observes the changes before and after the page execution at each GUI step and provides the agent with the reflected action history of the current step to avoid falling into an infinite backtracking loop.

\section{Prompt}\label{appendix_A}
Here, we give the prompt for action generation in Table \ref{prompt_action}, the prompt for the judgment module in Table \ref{prompt_judgment}, and the prompt for the reflection module in Table \ref{prompt_reflection}.

We fill in the prompt with the example in Figure \ref{figure3-1}. The inputs of the generator, judger and reflector are shown in below three tables respectively. As can be seen from the tables, the input of the generator needs to fill the current GUI page, the action space of the current GUI page, the history action list and the given task. The input of the judger still needs the action to be judged and the next page generated by the execution of that action based on the input of the generator. The input of the reflector still requires all the actions generated by multiple reflections and the next page generated by the execution of the last generated action based on the input of the generator.

\begin{tcolorbox}[title = {An example input of the generator.}, left = 0mm, breakable]

image\_path: .../Starbucks0\_10\_5\_2\_3\_6-screen.png

-----------------------------------------------------

The actions you can use are:

click("IngredientButton",[953,637][1068,752])

click("BackButton",[46,150][138,242])

click("StepperReduce",[790,1329][872,1411])

click("StepperAdd",[964,1329][1046,1411])

click("MediumCup",[44,1556][363,1820])

click("LargeCup",[382,1556][698,1820])

click("ExtraLargeCup",[717,1556][1036,1820])

click("Hot",[44,1930][1036,2059])

click("Ice",[382,1963][698,2059])

click("LightIce",[717,1963][1036,2059])

click("resetRecipe",[46,2126][362,2253])

click("addToCart",[385,2126][1034,2253])      

scroll("Customize",[0,1474][1080,2400],"up")

scroll("Customize",[0,1474][1080,2400],"down")

scroll("Customize",[0,1474][1080,2400],"left")

scroll("Customize",[0,1474][1080,2400],"right")

-----------------------------------------------------

You need to complete the following task:

I'd like to order a large cup of black tea latte, with extra Tahitian vanilla syrup, delivered to my home. 

-----------------------------------------------------

The completed actions are as follows:

click("delivery\_entry",[375,740][704,1032])

click("search",[530,748][783,841])

input("input",[46,242][848,346],"blact tea latte")

click("search",[894,230][1034,346])

click("addToCart",[953,709][1022,778])

\end{tcolorbox}

\begin{tcolorbox}[title = {An example input of the judger.}, left = 0mm, breakable]

image\_path:.../Starbucks0\_10\_5\_2\_3\_6-screen.png

-----------------------------------------------------

The actions you can use are:

click("IngredientButton",[953,637][1068,752])

click("BackButton",[46,150][138,242])

click("StepperReduce",[790,1329][872,1411])

click("StepperAdd",[964,1329][1046,1411])

click("MediumCup",[44,1556][363,1820])

click("LargeCup",[382,1556][698,1820])

click("ExtraLargeCup",[717,1556][1036,1820])

click("Hot",[44,1930][1036,2059])

click("Ice",[382,1963][698,2059])

click("LightIce",[717,1963][1036,2059])

click("resetRecipe",[46,2126][362,2253])

click("addToCart",[385,2126][1034,2253])      

scroll("Customize",[0,1474][1080,2400],"up")

scroll("Customize",[0,1474][1080,2400],"down")

scroll("Customize",[0,1474][1080,2400],"left")

scroll("Customize",[0,1474][1080,2400],"right")

-----------------------------------------------------

You need to complete the following task:

I'd like to order a large cup of black tea latte, with extra Tahitian vanilla syrup, delivered to my home. 

-----------------------------------------------------

The completed actions are as follows:

click("delivery\_entry",[375,740][704,1032])

click("search",[530,748][783,841])

input("input",[46,242][848,346],"blact tea latte")

click("search",[894,230][1034,346])

click("addToCart",[953,709][1022,778])

-----------------------------------------------------

Judgment: Please analyze whether the next action is helpful to further complete the task based on the current status and completed actions.

Next action: scroll(“Customize",[0,1474][1080,2400],“down") 

The page changes caused by executing the action are as follows:

image\_path:.../Starbucks0\_10\_5\_2\_3\_6-down-screen.png

-----------------------------------------------------

Final judgment (whether the next action is helpful to complete the task): (Yes or No)

\end{tcolorbox}
\begin{tcolorbox}[title = {An example input of the reflector.}, left = 0mm, breakable]

image\_path: .../Starbucks0\_10\_5\_2\_3\_6-screen.png

-----------------------------------------------------

The actions you can use are:

click("IngredientButton",[953,637][1068,752])

click("BackButton",[46,150][138,242])

click("StepperReduce",[790,1329][872,1411])

click("StepperAdd",[964,1329][1046,1411])

click("MediumCup",[44,1556][363,1820])

click("LargeCup",[382,1556][698,1820])

click("ExtraLargeCup",[717,1556][1036,1820])

click("Hot",[44,1930][1036,2059])

click("Ice",[382,1963][698,2059])

click("LightIce",[717,1963][1036,2059])

click("resetRecipe",[46,2126][362,2253])

click("addToCart",[385,2126][1034,2253])      

scroll("Customize",[0,1474][1080,2400],"up")

scroll("Customize",[0,1474][1080,2400],"down")

scroll("Customize",[0,1474][1080,2400],"left")

scroll("Customize",[0,1474][1080,2400],"right")

-----------------------------------------------------

You need to complete the following task:

I'd like to order a large cup of black tea latte, with extra Tahitian vanilla syrup, delivered to my home. 

-----------------------------------------------------

The completed actions are as follows:

click("delivery\_entry",[375,740][704,1032])

click("search",[530,748][783,841])

input("input",[46,242][848,346],"blact tea latte")

click("search",[894,230][1034,346])

click("addToCart",[953,709][1022,778])

-----------------------------------------------------

Reflection: This is not your first attempt to generate the next action. The previous attempts to generate the next action have all failed. 

Here are some previously generated next actions:

click("ExtraLargeCup",[717,1556][1036,1820])

click("StepperAdd",[964,1329][1046,1411])

scroll(“Customize",[0,1474][1080,2400],“down") 

-----------------------------------------------------

The page changes caused by executing the action are as follows:

image\_path: .../Starbucks0\_10\_5\_2\_3\_6-down-screen.png

Please note that you are currently in the middle stage of the trajectory. First, you need to analyze the current state, completed actions, and tasks, and compare them with the previous attempts at the next action. Then, you need to generate a new action that is different from all previously generated next actions.

\end{tcolorbox}

\newpage

\begin{figure*}[!t]
  \centering
  \includegraphics[width=1\textwidth]{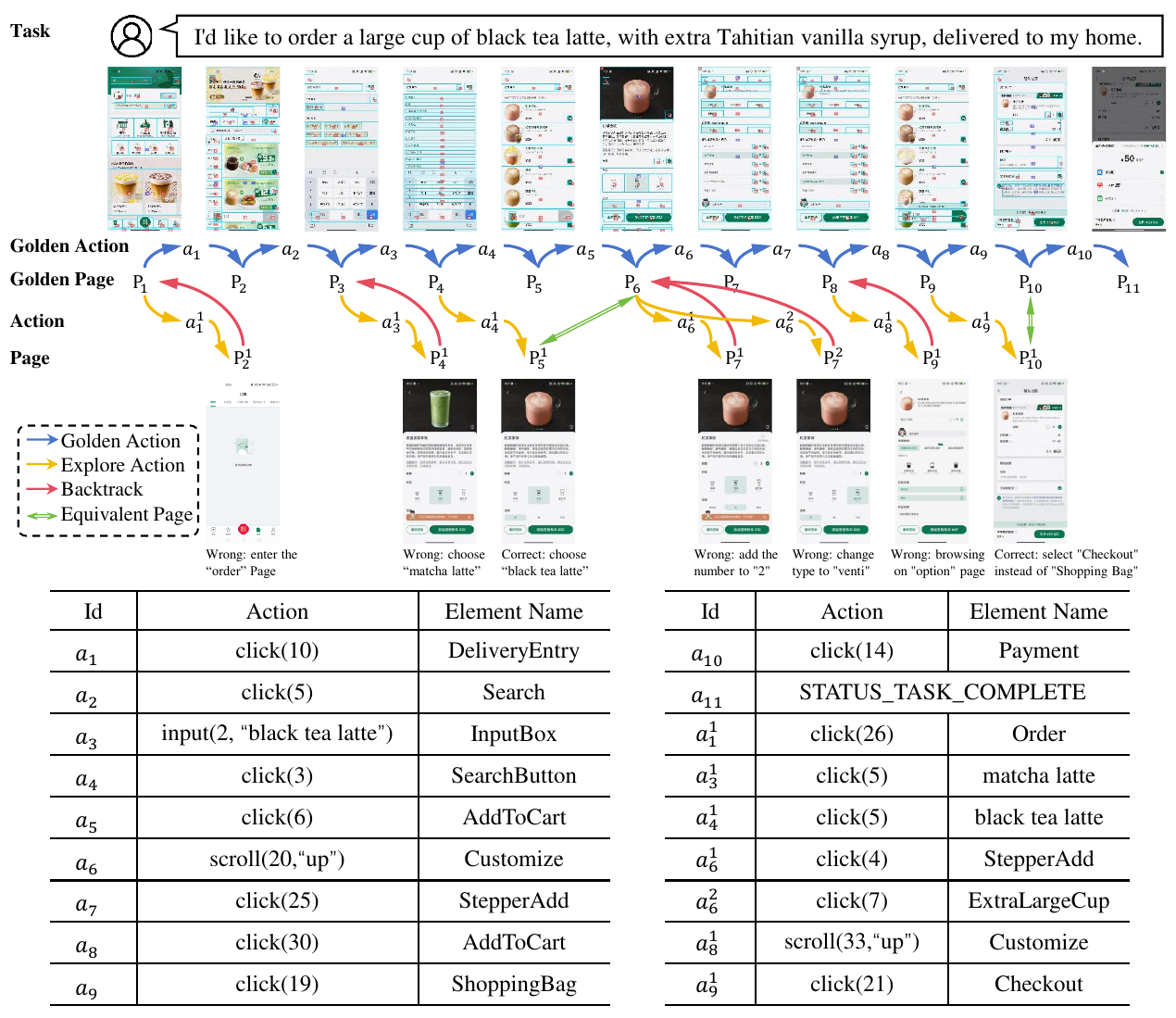}
  \caption{The complete 10-step GUI trajectory for a task. Green boxes represent the pages that need to be reached, and green circles represent the operations that need to be done. Orange arrows are the actions in the golden flow. Blue arrows are the actions in other GUI trajectories. Both the orange and blue flows can complete the task.}
  \label{figure2-2}
\end{figure*}

\section{The Judgment and Reflection Dataset Construction} \label{construct}

In this section, we introduce how to generate judgment and reflection datasets.
Taking a step from Figure \ref{figure2-2} as an example, the input and output of the original golden answer are:

\begin{tcolorbox}  

Input: X, $a_{<6}$, ActionSpace($\rm P_6$), $\rm P_6$. \\
Output: scroll ("Customize", "up") 

\end{tcolorbox}

Assume that the generator generates a new action, "click ("StepperAdd")", when regenerating this input. The evaluation index considers this action to be incorrect. Then, for the above two actions, we can construct two judgment data.

\begin{tcolorbox}  

****Case 1****

Input: X, $a_{<6}$, ActionSpace($\rm P_6$), $\rm P_6$. \\
Next action: scroll ("Customize", "up")  \\
Final judgment: (Yes or No) \\
Output: Yes 

---------------------------------------------------

****Case 2****

Input: X, $a_{<6}$, ActionSpace($\rm P_6$), $\rm P_6$. \\
Next action: click ("StepperAdd") \\
Final judgment: (Yes or No) \\
Output: No 

\end{tcolorbox}

Since the amount of data that does not require reflection is much larger than the data that needs reflection, we randomly select all negative data and 20\% of positive data to construct the reflection dataset. The reflection data formed by the above two judgment examples are as follows:

\begin{tcolorbox}  

****Case 1****

Input: X, $a_{<6}$, ActionSpace($\rm P_6$), $\rm P_6$. \\
Previous reflection list:  \\
scroll ("Customize", "up")  \\
You need to generate a new action. \\
Output: scroll ("Customize", "up") 

---------------------------------------------------

****Case 2****

Input: X, $a_{<6}$, ActionSpace($\rm P_6$), $\rm P_6$. \\
Previous reflection list:  \\
click ("StepperAdd")  \\
You need to generate a new action. \\
Output: scroll ("Customize", "up") 

\end{tcolorbox}

Here, during testing, the agent can perform multiple reflections until a satisfactory action is generated. When constructing the relection dataset, if the actions generated by the generator multiple times do not meet the evaluation metric, we will provide all of them to the reflector as a history reflection action list.

\begin{tcolorbox}  

Input: X, $a_{<6}$, ActionSpace($\rm P_6$), $\rm P_6$. \\
Previous reflection list:  \\
click ("StepperAdd")  \\
click ("ExtraLargeCup") \\
scroll ("Customize", "down") \\
You need to generate a new action. \\
Output: scroll ("Customize", "up") 

\end{tcolorbox}

\begin{figure*}[!t]
\centering
\includegraphics[width=1\textwidth]{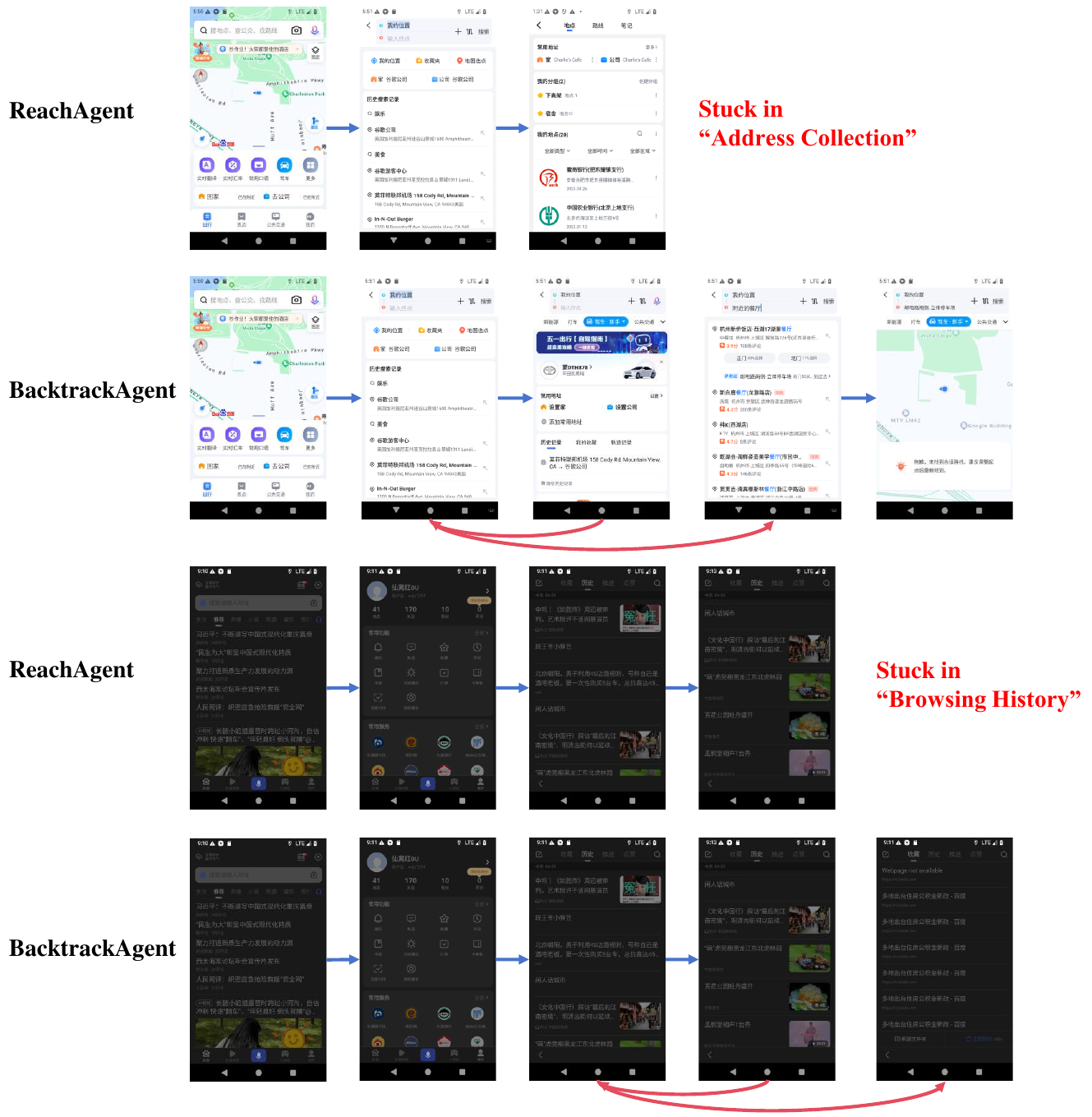}
\caption{Two cases of generated GUI chain by BacktrackAgent and ReachAgent.}
\label{figure6-1}
\end{figure*}
\section{A Step-by-Step Inference Process with Backtrack Mechanism}\label{step_by_step_inference}
Figure \ref{figure2-2} shows the complete action execution process of the GUI trajectory in Figure  \ref{figure2-1}. Here we provide a step-by-step reasoning process with backtracking for this example as follows:

1. On the Starbucks homepage, BacktrackAgent decides to click the Order button.

\begin{tcolorbox}  

$\rm P_1$ -> click ("Order") -> $\rm P_2^1$

\end{tcolorbox}

After observing the action execution result page, the error detection module found that the agent went to the order page without selecting coffee and decided to start the backtrack. 

The error recovery module reflects the action of the current step and decides to click the delivery entry button.

\begin{tcolorbox}  
$\rm P_1$ -> click ("DeliveryEntry") -> $\rm P_2$
\end{tcolorbox}

After discovering that the agent has entered the delivery entry page, the error detection module considers this action to be correct and decides to proceed to the next step.

2. On the delivery entry page, the generator decides to click the Search button.

\begin{tcolorbox}  
$\rm P_2$ -> click ("Search") -> $\rm P_3$
\end{tcolorbox}

The error detection module believes that entering the search page helps complete the task and proceeds to the next step.

3. On the search page, the agent decides to click the matcha latte button in the recommendation column.

\begin{tcolorbox}  
$\rm P_3$ -> click ("matcha latte") -> $\rm P_4^1$
\end{tcolorbox}

The error detection module finds that the agent has entered the product page of Matcha Latte and starts to backtrack. The reflector rewrites the current action to input the "black tea latte" in the search box.

\begin{tcolorbox}  
$\rm P_3$ -> input ("InputBox", "black tea latte") -> $\rm P_4$
\end{tcolorbox}

The error detection module adopts this action and goes to step 4.

4. After entering "black tea latte", the agent clicks the search button. The error detection module also considers this action to be correct.

\begin{tcolorbox}  
$\rm P_4$ -> click ("SearchButton") -> $\rm P_5$
\end{tcolorbox}

5. On the search results page for "black tea latte", the agent clicks the add button for the product. The error detection module decides to go directly to step 6.

\begin{tcolorbox}  
$\rm P_5$ -> click ("AddToCart") -> $\rm P_6$
\end{tcolorbox}

6. On the product page for “black tea latte,” the agent first clicks the plus icon in the number of cups.

The error detection module finds that the current action selects two cups of coffee when the task requires one. The reflector rewrites the current action and decides to select the extra-large cup.

BacktrackAgent finds that the extra-large cup is inconsistent with the task, and the action still needs to be rewritten. The agent chooses to slide up this time.

\begin{tcolorbox}  

$\rm P_6$ -> click ("StepperAdd") -> $\rm P_7^1$

---------------------------------------------------

$\rm P_6$ -> click ("ExtraLargeCup") -> $\rm P_7^2$

---------------------------------------------------

$\rm P_6$ -> scroll ("Customize", "up") -> $\rm P_7$
\end{tcolorbox}

The agent confirms that there are no parameters on the previous product page that need to be modified by the agent. It needs to browse the parameter page to find the new parameters. The action is correct, go to step 7.

7. On the parameter page, the BacktrackAgent clicks the Add button for Tahitian vanilla syrup. The error detection module passes this action.

\begin{tcolorbox}  
$\rm P_7$ -> click ("StepperAdd") -> $\rm P_8$
\end{tcolorbox}

8. After selecting the parameters, the agent decides to continue swiping up to browse more parameters.

The error detection module finds that all parameters of the "black tea latte" have been customized and there is no need to continue browsing. The error recovery module changes the current action to add to the shopping cart.

\begin{tcolorbox}  

$\rm P_8$ -> scroll ("Customize", "up")  -> $\rm P_9^1$

---------------------------------------------------

$\rm P_8$ -> click ("AddToCart")  -> $\rm P_9$
\end{tcolorbox}

9. BacktrackAgent decides to click the shopping bag button. The error detection module sees that it has reached the checkout page and goes to step 10.

\begin{tcolorbox}  
$\rm P_9$ -> click ("ShoppingBag")  -> $\rm P_{10}$
\end{tcolorbox}

10. BacktrackAgent clicks the payment button. The error detection module passes this action.

\begin{tcolorbox}  
$\rm P_{10}$ -> click ("Payment")  -> $\rm P_{11}$
\end{tcolorbox}

11. BacktrackAgent believes that the task has been completed and generates the special token "STATUS\_TASK\_COMPLETE" to end the reasoning process.
\begin{tcolorbox}  
$\rm P_{11}$ -> STATUS\_TASK\_COMPLETE
\end{tcolorbox}

\section{Case Study} \label{case}
Here we provide two cases of errors during evaluation (See Figure \ref{figure6-1}). 
We can see that the ReachAgent predicts several steps correctly but if one action is wrong, the agent would fail the task. In contrast, when BacktrackAgent mistakenly enters the "Address Collection" page and browses on the "Browsing History", it can detect the error and recover to the correct track, and finally complete the task.

The step-by-step case is described as follows, given the task "Find a route to a nearby restaurant.", agent observes that the current page is the home page of BaiduMap APP. 

In step 1, the agent successfully clicked on the route search and navigated to the search page. 

In the step 2, the agent first mistakenly went to the collection page. The error detection module discovered the error, and the error recovery module revised the action to the input "nearby restaurants" and navigated to the search results page. 

In the step 3, the agent clicks on a restaurant and gets the route to that restaurant.

In step 4, the agent considers the task completed and decides to exit.

\begin{tcolorbox}  

Step 1:

Action: click("Route Search")
    
**Execute the Action**

Observation: The search page

Error Detection: Correct

---------------------------------------------------

Step 2:

Action: click("Address Collection")

**Execute the Action**

Observation: The collection page

Error Detection: Wrong

Error Recovery: input("Endpoint", "nearby restaurant")

**Execute the Action**

Observation: The search result page of nearby restaurant

Error Detection: Correct

---------------------------------------------------

Step 3:

Action: click("Parking Lot")

**Execute the Action**

Observation: The route to the Lake View Restaurant's parking lot

Error Detection: Correct

---------------------------------------------------

Step 4:

Action: Complete

\end{tcolorbox}

\begin{table*}[!t]
    \centering
\setlength{\abovecaptionskip}{0.3cm}
\resizebox{1\textwidth}{!}{
    \begin{tabular}{l|cccccc|cccccc}
    \hline\hline
    \multirow{2}{*}{Model}  &  \multicolumn{6}{c|}{Task Level}  &  \multicolumn{6}{c}{Step Level}   \\ \cline{2-13} 
        ~ & \underline{Overall}  & General & Google & Install & Single & WebShop & \underline{Overall} & General & Google & Install & Single & WebShop\\ \hline
        GPT-4o & 15.16 & 1.1  & 	10.73 & 	3.03 & 	46.81 & 	7.96 &55.38 & 47.06 & 52.30 & 49.12 & 80.28 & 46.42 \\ 
        Auto-UI$\rm _{unified}$ &- & - & - & - & - & - & 74.52& 68.24 & 71.37 & 76.89 & 84.58 & 70.26 \\ 
        Auto-UI$\rm _{separate}$ & - & - & - & - & - & - & 75.13& 65.94 & 76.45 & 77.62 & 81.39 & 69.72 \\ \hline
        Qwen-VL & 16.97 & 12.28 & 12.96 & 17.32 & 38.17 & 6.35 &68.75 & 62.11 & 67.13 & 75.68 & 73.08 & 64.12 \\ 
        Qwen2-VL  &21.56 & 16.51 & 13.16 & 22.53 & 47.89 & 11.31 & 72.26 & 67.50 & 67.70 & 78.45 & 76.54 & 70.74 \\ 
        MobileVLM$\rm _{unified}$ & 25.07 & 18.31 & 24.68 & 23.19 & 45.95 & 12.99 &75.81  & 69.58 & 74.72 & 79.87 & 81.24 & 71.70 \\ 
        MobileVLM$\rm _{separate}$ & 25.53 & 19.68 & \textbf{25.39} & 22.80 & 47.14 & 13.02 & 77.36 & 70.26 & \textbf{76.86} & 78.86 & \textbf{87.06} & 71.42 \\ 
        ReachAgent$\rm _{unified}$ & 24.89 & 19.89 & 24.74 & 23.89 & 48.70 & 9.38 & 74.54 & 70.27 & 74.94 & 80.76 & 77.17 & 69.02 \\ 
        ReachAgent$\rm _{separate}$ & 25.28 & 22.22 & 24.68 & 22.61 & 46.26 & 12.99 & 74.81 & 70.16 & 74.86 & 79.41 & 76.26 & 71.70 \\ \hline
        Generator & 25.83& 18.10 &	20.87	& 26.07 &	53.29 &	12.38 &	75.36&	68.85 &	73.16 &	80.61 &	80.44 &	71.88 \\
        BacktrackAgent & \textbf{29.72} & \textbf{22.60} & 23.46 & \textbf{27.33} & \textbf{58.79} & \textbf{15.14} & \textbf{78.04} & \textbf{71.58} & 75.75 & \textbf{82.11} & 82.61 & \textbf{74.78} \\ \hline\hline
    \end{tabular}}
    \caption{Main Result(\%) on AutoUI dataset. "separate" means that this baseline is trained on five subsets of Auto-UI, while "unified" means that the baseline is trained on the entire Auto-UI dataset as a whole. - means that Auto-UI only provides official accuracy at the step level, and does not provide test results for us to calculate the accuracy at the task level.
    }
    \label{result_2}
\end{table*}

\section{Additional Experiment of Main Results} 
\label{appendix_autoui}
This section provides detailed experimental data on Auto-UI. From Table \ref{result_2} we can see,

$\bullet$  BacktrackAgent outperforms the SOTA baseline in General, Install, Single, and WebShop splits. Unlike MobileVLM, we did not pre-train Qwen2-VL with additional data to achieve this result. In addition, we only used simulated execution pages to train the Judger and Reflector. The improvement of BacktrackAgent over Qwen2-VL and Generator proves that our framework and backtracking mechanism can generally improve task completion abilities on different datasets.

$\bullet$   BacktrackAgent performs slightly better than the SOTA baseline in General, Install and WebShop splits, and slightly worse in Google and Single splits.

$\bullet$  Compared with the backbone model Qwen2-VL, BacktrackAgent significantly outperforms it in every split. This proves the effectiveness of our framework and backtracking mechanism.

The above experimental results show that BacktrackAgent achieves comparable results to the SOTA Agent MobileVLM, while BacktrackAgent is still significantly better than our baseline Qwen2-VL.  This is because:

$\bullet$  Unlike MobileVLM, we did not use additional data to pre-train Qwen2-VL.

$\bullet$ We only used simulated execution pages to train Judger and Reflector, since Auto-UI did not provide actual execution results. The ablation experimental table above also verifies that the actual execution strategy is significantly better than the simulated execution strategy.

\begin{table*}[ht]
    \centering

\resizebox{1\textwidth}{!}{
    \begin{tabular}{ll|c|ccc|cc}
    \hline\hline
        \multirow{2}{*}{Model}  & \multirow{2}{*}{Method}   &  Task Success & \multicolumn{3}{c|}{Task Level Acc}  &  \multicolumn{2}{c}{Step Level Acc}   \\  \cline{4-8} 
        & &  Rate & Both &  IoU & Text & IoU  & Text   \\ \hline 
        ReachAgent	 &SFT	 &45.33	 &27.48	 &37.82	 &31.31	 &83.34	 &81.47 \\ 
ReachAgent	 &SFT+RL	 &46.52	 &29.79	 &38.75	 &33.06	 &83.32	 &81.77 \\  \hline 
\textbf{BacktrackAgent}	 &Original Agent	 &\textbf{54.11} & \textbf{33.51} & \textbf{43.25} & \textbf{36.67} & \textbf{84.94} & \textbf{83.24} \\ 
$\Delta$	 &	 &+7.59	 &+3.72	 &+4.50	 &+3.61	 &+1.62	 &+1.47 \\ \hline\hline
\multicolumn{8}{c}{10 repeated tests, each with 80\% of the test dataset} \\  \hline
Repetition 1	 &	 &54.30	 &33.19	 &43.14	 &36.45	 &84.94	 &83.21\\  
Repetition 2	 &	 &54.21	 &33.84	 &43.61	 &37.19	 &84.95	 &83.24\\  
Repetition 3	 &	 &53.23	 &32.96	 &42.72	 &36.45	 &84.80	 &83.12\\  
Repetition 4	 &	 &55.32	 &33.66	 &43.70	 &37.05	 &85.04	 &83.33\\  
Repetition 5	 &	 &53.74	 &33.33	 &42.72	 &36.68	 &84.78	 &83.15\\  
Repetition 6	 &	 &54.07	 &33.84	 &43.84	 &36.91	 &85.11	 &83.36\\  
Repetition 7	 &	 &54.77	 &34.31	 &43.56	 &37.38	 &85.03	 &83.44\\  
Repetition 8	 &	 &54.16	 &33.24	 &43.00	 &36.68	 &84.88	 &83.25\\  
Repetition 9	 &	 &53.84	 &33.10	 &43.10	 &36.12	 &84.92	 &83.13\\  
Repetition 10	 &	 &53.60	 &33.01	 &42.63	 &36.12	 &84.79	 &83.09\\   \hline
\textbf{BacktrackAgent}	 & 11 Tests' Avg.  &\textbf{54.12±0.57}	 &\textbf{33.45±0.42}	 &\textbf{43.21±0.42}	 &\textbf{36.70±0.41}	 &\textbf{84.93±0.11}	 &\textbf{83.23±0.11}\\  
$\Delta$	 &	 &+7.60	 &+3.66	 &+4.46	 &+3.64	 &+1.61	 &+1.46\\  
p-value	 &	 &4.119e-19	 &9.425e-16	 &3.363e-17	 &5.997e-16	 &6.897e-20	 &4.768e-19\\  \hline\hline
\multicolumn{8}{c}{ 2 additional BacktrackAgents trained with different seeds }\\    \hline
Retrained 1	 &	 &53.7	 &32.73	 &42.51	 &36.18	 &84.87	 &83.14\\  
Retrained 2	 &	 &53.85	 &33.25	 &43.4	 &36.67	 &85.11	 &83.31\\    \hline
\textbf{BacktrackAgent}	 & 	3 Agents' Avg.  &\textbf{53.89±0.21}	 &\textbf{33.16±0.40}	 &\textbf{43.05±0.48}	 &\textbf{36.51±0.28}	 &\textbf{84.97±0.12}	 &\textbf{83.23±0.09}\\  
$\Delta$	 &	 &+7.37	 &+3.37	 &+4.30	 &+3.45	 &+1.65	 &+1.46\\  
p-value	 &	 &4.184e-7	 &1.242e-4	 &9.752e-5	 &2.981e-5	 &2.045e-5	 &7.759e-6 \\ \hline\hline
    \end{tabular}}
    \caption{Main Result(\%) on Mobile3M dataset. The top part is the results of the SOTA model ReachAgent, which also includes two-stage SFT and RL. The middle part is the evaluation results of 10 samplings on the test set using the original BacktrackAgent. The bottom is 2 additional BacktrackAgents trained with different seeds, which are used together with the original BacktrackAgent to calculate the overall performance. The overall evaluation metric is mean±SD (54.12±0.57), where mean represents the mean of multiple tests and SD represents the standard deviation. $\Delta$ represents the difference between the mean and ReachAgent, i.e., the performance improvement. The p-value is calculated using T-test \cite{student1908probable}.
    }
    \label{result_significant}
\end{table*}

\section{Statistically Significant Experiment} 
\label{appendix_significant}

To ensure that we observed statistically significant differences between BacktrackAgent and other SOTA Agents, we performed statistical significance tests. 

$\bullet$  For the test data, we conducted 10 repeated experiments on the test set, randomly sampling 80\% of the test examples each time. 

$\bullet$  For BacktrackAgent itself, we retrained the agent twice from the backbone model with different seeds, using the same training data and model parameters as BacktrackAgent, including stage 1 SFT and stage 2 RL. 

We conducted these experiments to verify that BacktrackAgent has statistically significant performance differences compared to SOTA Agents. The experimental results are shown in the Table \ref{result_significant}. From the table, we can see that:

$\bullet$ The 10 experiments sampled on the test set have high significant p-values (p < 5.9*10-16) on all evaluation metrics, confirming that there is a significant difference in performance between BacktrackAgent and the SOTA model.

$\bullet$ The two versions of BacktrackAgent retrained with different seeds, together with the original BacktrackAgent, also achieved significant p-values (p < 1.3*10-4) on all evaluation metrics, which also proves that the agent performance is significant and reproducible.

$\bullet$ From the observation data, it can be seen that for the Task Success Rate, BacktrackAgent has achieved a 7.59\% improvement over ReachAgent, and the fluctuation caused by both the resampled test set and the retrained Agent on performance is less than 1.2\%. This is enough to prove that the performance achieved by BacktrackAgent is statistically significant and has good stability. Similarly, for the Step-level Accuracy, the fluctuation caused by different repetitions is less than 0.2\% when the agent achieves a performance improvement of 1.62\% and 1.47\%. This is because the step-level accuracy itself exceeds 80\%, and there is little room for improvement. But the agent's performance on this metric is also stable. 

$\bullet$ The consistency across repetitions also shows that our improvements are reliable and not random.

\end{document}